\documentclass[twocolumn]{article}
\usepackage{graphicx} 
\usepackage[margin=1in]{geometry} 
\usepackage{xcolor} 
\usepackage[normalem]{ulem} 
\usepackage[pagewise]{lineno} 
\usepackage{amsmath} 
\usepackage{amssymb} 
\usepackage{bm} 
\usepackage{empheq}
\usepackage{abstract}
\usepackage{optidef}
\usepackage{float}
\usepackage{stfloats}
\usepackage{placeins}
\usepackage{authblk} 

\newcommand*{\bl}{}

\date{}
\begin{document}

\title{ATMO: An  Aerially Transforming Morphobot for Dynamic Ground-Aerial Transition}

\author[1,2]{Ioannis Mandralis$^\ast$}
\author[1]{Reza Nemovi}
\author[3]{Alireza Ramezani}
\author[2]{Richard M. Murray}
\author[1]{Morteza Gharib\small}
\affil[1]{Department of Aerospace Engineering, California Institute of Technology}
\affil[2]{Department of Computing and Mathematical Sciences, California Institute of Technology}
\affil[3]{Electrical and Computer Engineering Department, Northeastern University}

\twocolumn[
\maketitle
\vspace{-4em}
\begin{center}
{\small $^{\ast}$Corresponding author: imandralis@caltech.edu}
\end{center}
\begin{abstract}
Designing ground-aerial robots is challenging due to the increased actuation requirements which can lead to added weight and reduced locomotion efficiency. Morphobots mitigate this by combining actuators into multi-functional groups and leveraging ground transformation to achieve different locomotion modes. However, transforming on the ground requires dealing with the complexity of ground-vehicle interactions during morphing, limiting applicability on rough terrain. Mid-air transformation offers a solution to this issue but demands operating near or beyond actuator limits while managing complex aerodynamic forces. We address this problem by introducing the Aerially Transforming Morphobot (ATMO), a robot which transforms near the ground achieving smooth transition between aerial and ground modes. To achieve this, we leverage the near ground aerodynamics, uncovered by experimental load cell testing, and stabilize the system using a model-predictive controller that adapts to ground proximity and body shape. The system is validated through numerous experimental demonstrations. We find that ATMO can land smoothly at body postures past its actuator saturation limits by virtue of the uncovered ground-effect.
\end{abstract}
]

\section*{Introduction}

\begin{figure*}[t]
    \centering
    \includegraphics[width=1\linewidth]{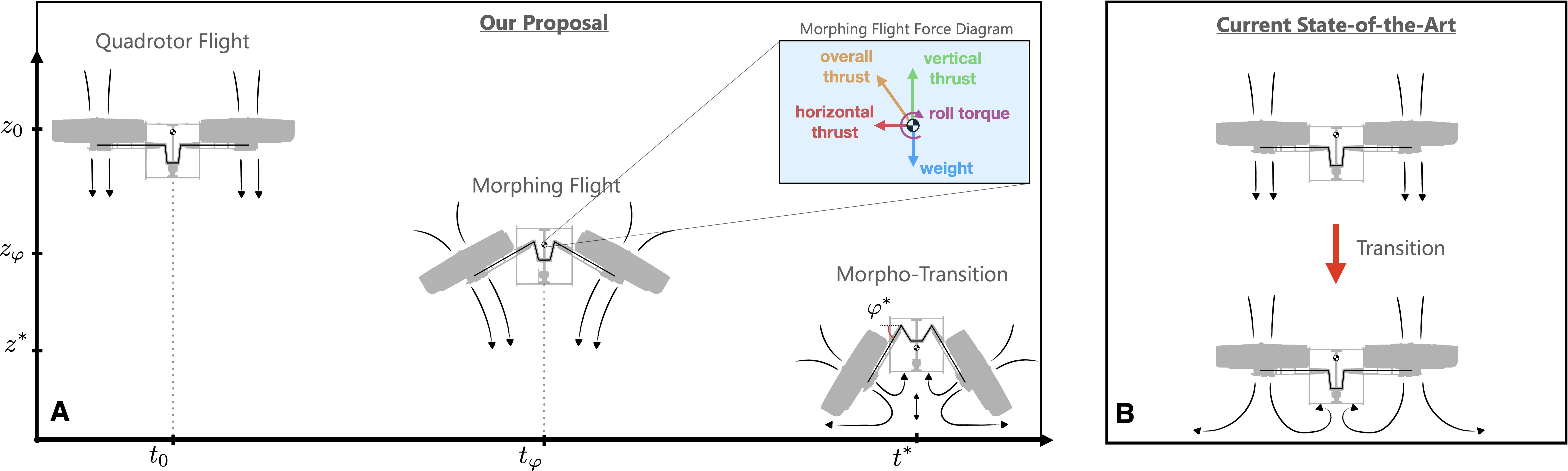}
    \caption{\textbf{Proposed Dynamic Wheel Landing Maneuver Versus Quadrotor Landing.} (\textbf{A}) Phases of the envisioned dynamic wheel landing maneuver classified according to altitude $z(t)$. The robot is initially in regular flight mode at some height $z_0$, as it descends it begins tilting its wheel-thrusters at some height $z_\varphi$. Finally, at some height $z^\ast$ the robot switches to the near ground controller and lands while maximizing the tilt angle $\varphi$. The flow field has been sketched to indicate the complexity of the aerodynamics at each phase. A free body diagram of the possible forces during morphing flight when the robot tries to compensate for roll disturbances is depicted in the top right corner. (\textbf{B}) Quadrotor transitioning to the ground and entering ground effect. The flow pattern is sketched qualitatively. }
    \label{fig:1}
\end{figure*}

Effective ground-aerial locomotion is important for a wide range of robotic applications, including last mile delivery \cite{Grubesic2024,Lemardel2021} or space exploration \cite{Ayele2022,Young2005}. However, both ground and aerial robots are still not able to reliably operate in the real world. Ground robots are limited by their range of operation making it impossible to traverse high obstacles or perform inspection tasks at an altitude whilst aerial robots have limited battery performance due to payload requirements as well as safety concerns when flying close to urban centers. Predictions on the future of autonomous robotic systems point to the fact that many challenges facing today's autonomous systems may be relieved if we combine aerial and terrestrial capabilities \cite{Floreano2015, Ramirez2023}, making the development of effective ground-aerial robots particularly pressing.

Many ground-aerial robot designs rely on the philosophy of redundancy and fulfill their bi-modal locomotion requirements using multiple actuators that can perform one function only \cite{weakly2024bistable,Zhang2021,Wang2021,Tanaka2017,Meiri2019,Kalantari2020,Choi2021}. However, these types of redundant robot designs often result in using more actuators and components than strictly necessary, increasing system weight and cost.  

Instead, Morphobots, or robots that reuse the same appendages {\bl for different tasks through shape change}, are able to generate different locomotion modes while reducing system weight and complexity \cite{Sihite2023}. These robot designs, often inspired by the multi-functional locomotion behaviors of animals in nature, are generally thought to enhance the efficiency of mobile autonomous robots faced with changing, unstructured environments \cite{baines2022multi,sun2023embedded,baines2024robots,zheng2020,Ruiz2022}. Indeed, recent work showed that using multi-functional appendages combined with body shape change resulted in increased locomotion plasticity, enabling maneuvers that were previously impossible \cite{Sihite2023}. 

A lesser explored capability of Morphobots is the ability to use shape change to transform mid-air as a means to enhance ground-aerial locomotion. This could provide them with the crucial ability to {\bl avoid ground-vehicle interaction during morphing}. For example, ground transformation may be impossible due to rough terrain that hinders the ground motion of the robot appendages. In such scenarios mid-air transformation may offer a reliable path to mission safety and behavioral agility. 

In this work we envision an aerial transition maneuver that links flying and driving locomotion called the dynamic wheel landing, depicted in Figure 1(A). In a dynamic wheel landing, the aim is to transition smoothly from flying to driving locomotion by morphing near the ground and landing on dual-purpose wheel-thruster appendages with the largest possible tilt angle, i.e. as close as possible to drive configuration, while achieving a desired impact velocity. In conventional quadrotor landing maneuvers, illustrated in Figure 1(B), the robot usually lands by vertical, non-transforming descent \cite{gautam2014}. Instead, the proposed maneuver involves morpho-transitioning i.e. transitioning between two modes of locomotion through near-ground morphing.

Achieving such a maneuver is challenging on multiple fronts. From a design perspective, the need to transform mid-air requires elevated torque to continually fight against the thrust forces. From a modeling and control perspective, mid-air morphing introduces new dynamic couplings between the robot degrees of freedom and actuator limits become increasingly critical, {\bl while autonomous near ground aerial operation is a known challenging problem due to ground effect aerodynamics \cite{Shi2019,Nonaka2011,Danjun2015,habibnia2019ann}.} Compounded with the fact that the aerodynamics of morphing-flight, as well as near ground transformation aerodynamics are mostly unknown, novel controllers, models, and design are required. 

We address the challenge of aerial transformation for Morphobots concretely by 1) designing a flying-driving Morphobot called the Aerially Transforming Morphobot (ATMO) that is specialized for mid-air transformation through a {\bl morphing mechanism} that enables body shape change mid-flight with minimal actuation requirements, 2) studying its physics to reveal important dynamic and aerodynamic dependencies, 3) using these findings to develop a general purpose ground-aerial morpho-transition controller, and 4) demonstrating the mid-air transformation capabilities experimentally. Our results show that using mid-air robotic transformation can result in dynamic ground-aerial transition maneuvers that enhance robot agility and expand operational range - paving the way for greater autonomy in future mobile robotic missions.

\section*{Results}

\begin{figure*}[t]
    \centering
    \includegraphics[width=1\linewidth]{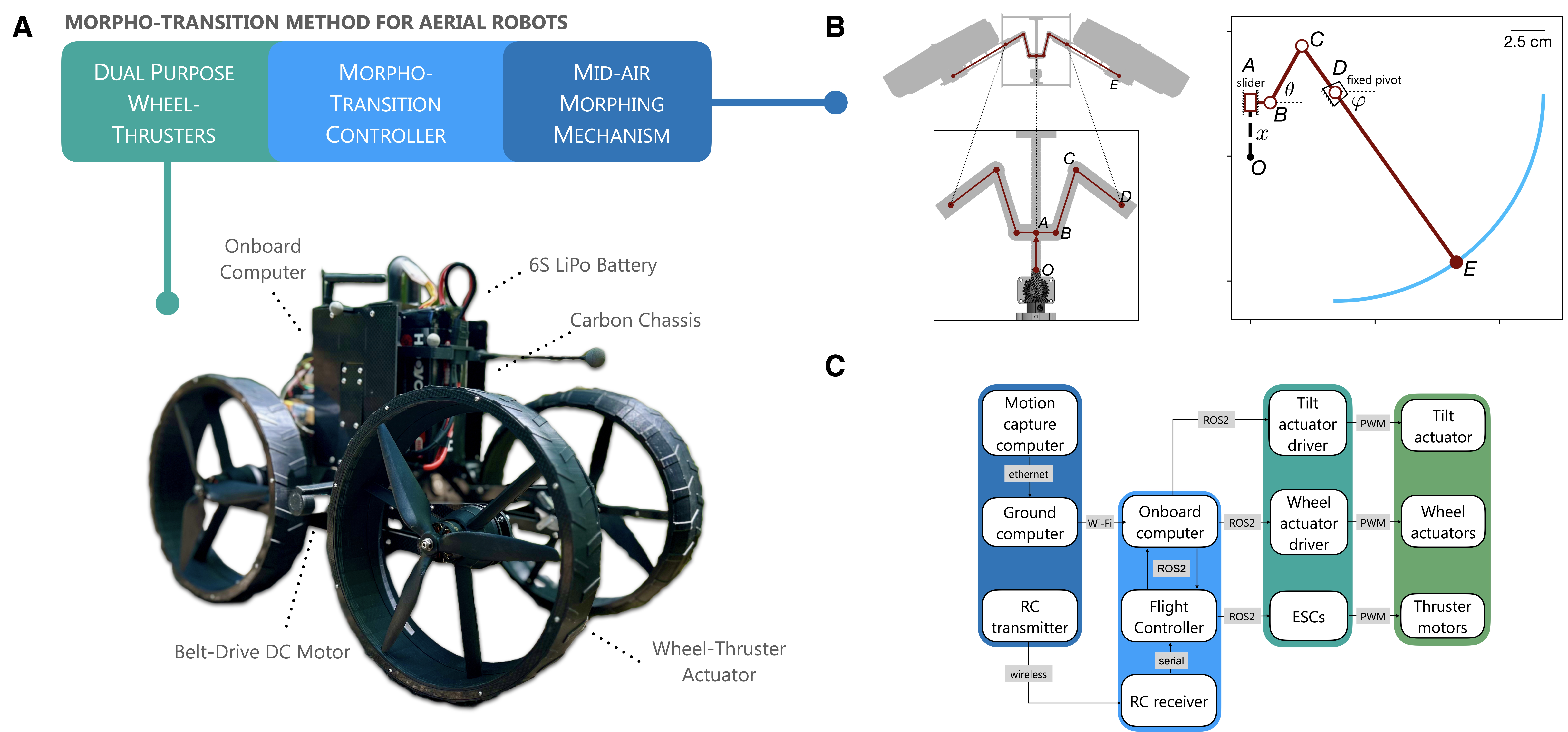}
    \caption{\textbf{Aerially Transforming Morphobot (ATMO) Overview} (\textbf{A}) Illustration of ATMO in perspective with key components labeled. (\textbf{B}) Tilt mechanism. (left) ATMO is shown in a morphed configuration in flight. Underneath, the tilt actuator box is enlarged and the joints are labeled. The tilt mechanism is actuated by two co-rotating bevel gears actuated by a DC motor. The spinning causes joint $A$ to translate on the $OA$ axis, inducing mechanism motion. (right) The right half of the symmetric mechanism is shown with all the joints labeled as well as point $E$ {\bl which corresponds to the intersection of the propeller axis with the extension of $CD$}. The path taken by joint $E$ as joint $A$ moves from bottom to top is traced in blue. {\bl Open colored joints represent rotating joints or sliders. $CDE$ is a unified link pivoted on D.} The key kinematics parameter is the tilt angle $\varphi$ which varies from $\varphi=0$ in flight configuration to $\varphi=\frac{\pi}{2}$ in drive configuration. (\textbf{C}) Electronics architecture and connections. On the left are ground control components. These communicate to the onboard components by Wi-Fi or wireless (radio protocol) transmission. The onboard computer receives radio signals from the flight controller and communicates to all the motor drivers using ROS2.}
    \label{fig:2}
\end{figure*}

\subsection*{Robotic Platform Overview}
The Aerially Transforming Morphobot (ATMO), depicted in Figure 2(A), was specially designed to solve the problem of mid-air transformation. To achieve this, we re-used some design principles from the M4 robot \cite{Sihite2023} with the key difference that all posture manipulation actuators were removed and replaced by a single morphing mechanism. The morphing mechanism, depicted in Figure 2(B) dynamically controls the tilt angle, $\varphi$, of four wheel-thruster pairs using a single brushed DC motor. It works using a closed-loop kinematic linkage that is actuated by a DC motor that rotates a central worm gear. This rotation results in linear motion of joint A which is then converted to rotational motion of joints B, C, and D - tilting the wheel-thruster appendages symmetrically. By virtue of the high reduction of the worm gear mechanism, the posture of the robot locks in place even when the tilt motor is not being actuated. This allows ATMO to fly in morphed configurations while avoiding actuator damage which could result in sudden loss of desired posture and flight failure. For an animation of the tilt actuator in action, the reader is referred to Video 9. Compared to the M4 robot \cite{Sihite2023}, the number of actuators controlling the body posture in order to achieve both flight and driving is reduced from 12 to 1 resulting in an overall simpler system with fewer failure points.

ATMO achieves flying and driving using its dual-purpose appendages through shape change. In flight mode, ATMO is configured as a conventional quadcopter and uses its wheel-thruster appendages for propulsion, whilst in drive mode the same appendages are reused for wheeled locomotion. The final result is a compact robot with total weight 5.5 kg, including the battery. The robot stands 16 cm tall and 65 cm wide in aerial configuration, and 33 cm tall and 30 cm wide in ground configuration. Driving is achieved by two belt-pulley systems on either side of the robot which are actuated by driving motors, enabling differential drive steering.  ATMO is equipped with an onboard computer running a custom controller, as well as onboard sensors for state estimation and fusion. All communication is achieved through the state-of-the-art message-passing middleware software ROS2 \cite{Macenski2022}. The electro-mechanical architecture and avionics are depicted in Figure 2(C). For a detailed breakdown of all of the robot subsystems, including the wheels and wheel drive system, the propellers and propulsion system, and the morphing mechanism, the reader is referred to the Supplementary Materials.

ATMO, has a thrust-to-weight ratio of 2.1:1 resulting in a critical tilt angle of $\varphi_c = 60^\circ$ (see Methods). In this configuration, the thruster actuators become fully saturated when trying to counteract gravity, and all available control authority is reduced to zero. Thus, minute disturbances are enough to destabilize the system, resulting in failure since the robot cannot hold its vertical position and compensate for disturbances simultaneously. The critical posture is thus the most extreme configuration with which the robot can descend and impact the ground with zero impact velocity, making it a key quantity to consider when performing dynamic ground-aerial transitions. In this work we have chosen to limit the in-flight body posture to $\varphi_{\max} = 50^\circ$, when not in transition close to the ground, ensuring that at least $35\%$ of the total thrust is always available for disturbance rejection. The reader is referred to the Methods section for further details. For a summary video of the robot's capabilities, the reader is referred to Video 1.

\begin{figure*}[!t]
    \centering
    \includegraphics[width=1\linewidth]{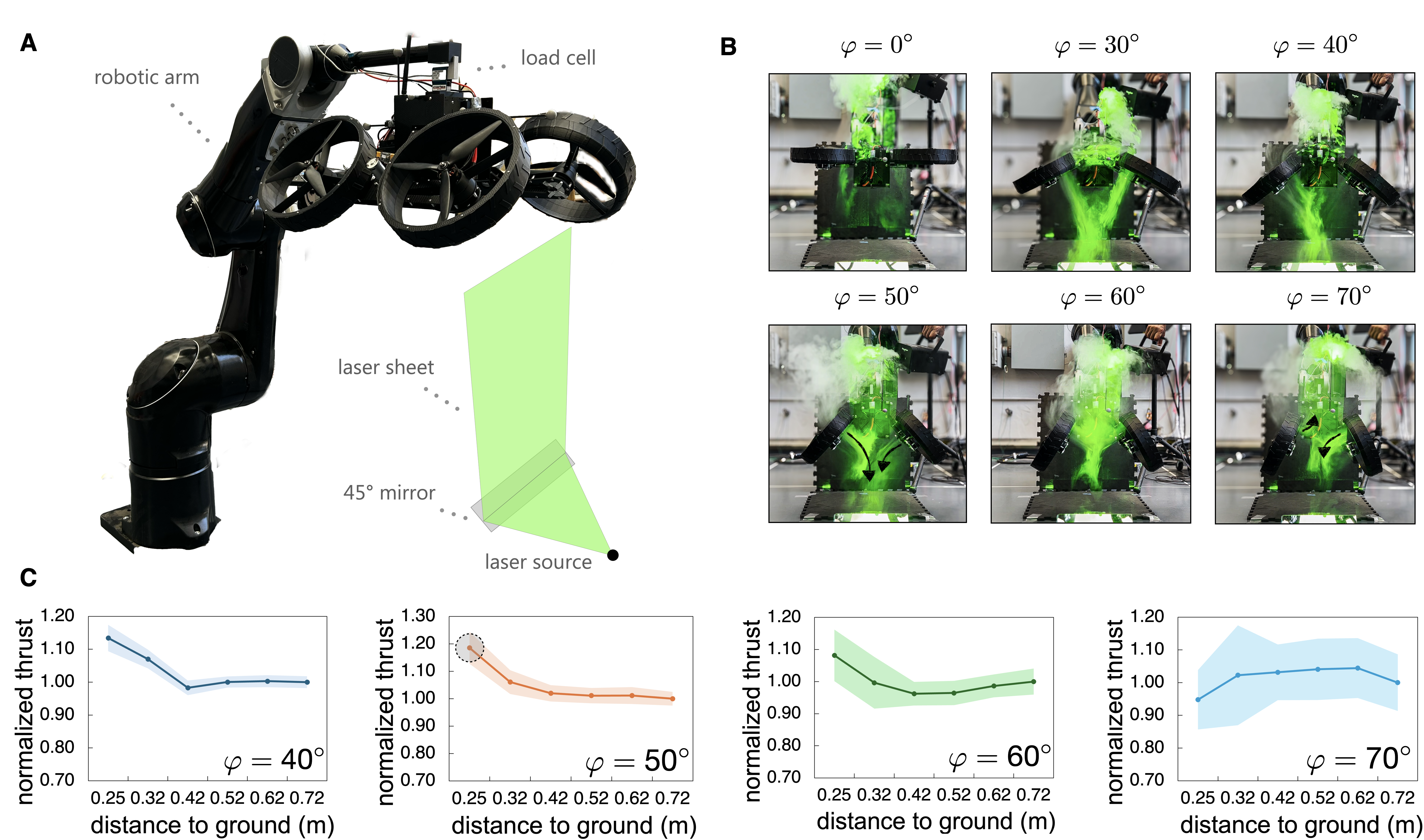}
    \caption{\textbf{Morpho-Transition Aerodynamics.} \textbf{(A)} Experimental setup. A 6 axis robotic arm is used to adjust the three dimensional position of ATMO. A load cell is in series between the robotic arm and ATMO. A laser sheet is positioned underneath the plane of the front two thrusters for imaging purposes. To achieve this a laser source is shone on a 45 degree mirror generating a vertical sheet which constitutes the imaging plane. (\textbf{B}) Smoke visualizations of the aerodynamic flow field with the robot stationary at tilt angles. For $\varphi=0^\circ$ two streams of air are flowing vertically through the wheel-thrusters. For $\varphi=30^\circ$ the two jets are reoriented and mix to form one downward oriented stream. At $\varphi=70^\circ$ the two jets impinge to form a stagnation point with an unstable region where the flow may be directed either upwards or downwards. Arrows indicated the overall direction of the flow as observed in the videos of the visualization (see supplementary materials). \textbf{(C)} Results from the load cell testing for $\varphi=40^\circ,50^\circ,60^\circ,70^\circ$. The overall thrust (measured at the load cell) normalized by the thrust produced in the same configuration far from the ground is plotted at 6 different heights $z=0.25,0.32,0.42,0.52,0.62,0.72 \textrm{m}$. The standard deviation of the thrust during the measured time horizon is shaded in lighter color for each experiment.}
    \label{fig:3}
\end{figure*}
\subsection*{Aerodynamic Characterization}

\begin{figure}[!t]
    \centering
    \includegraphics[width=\linewidth]{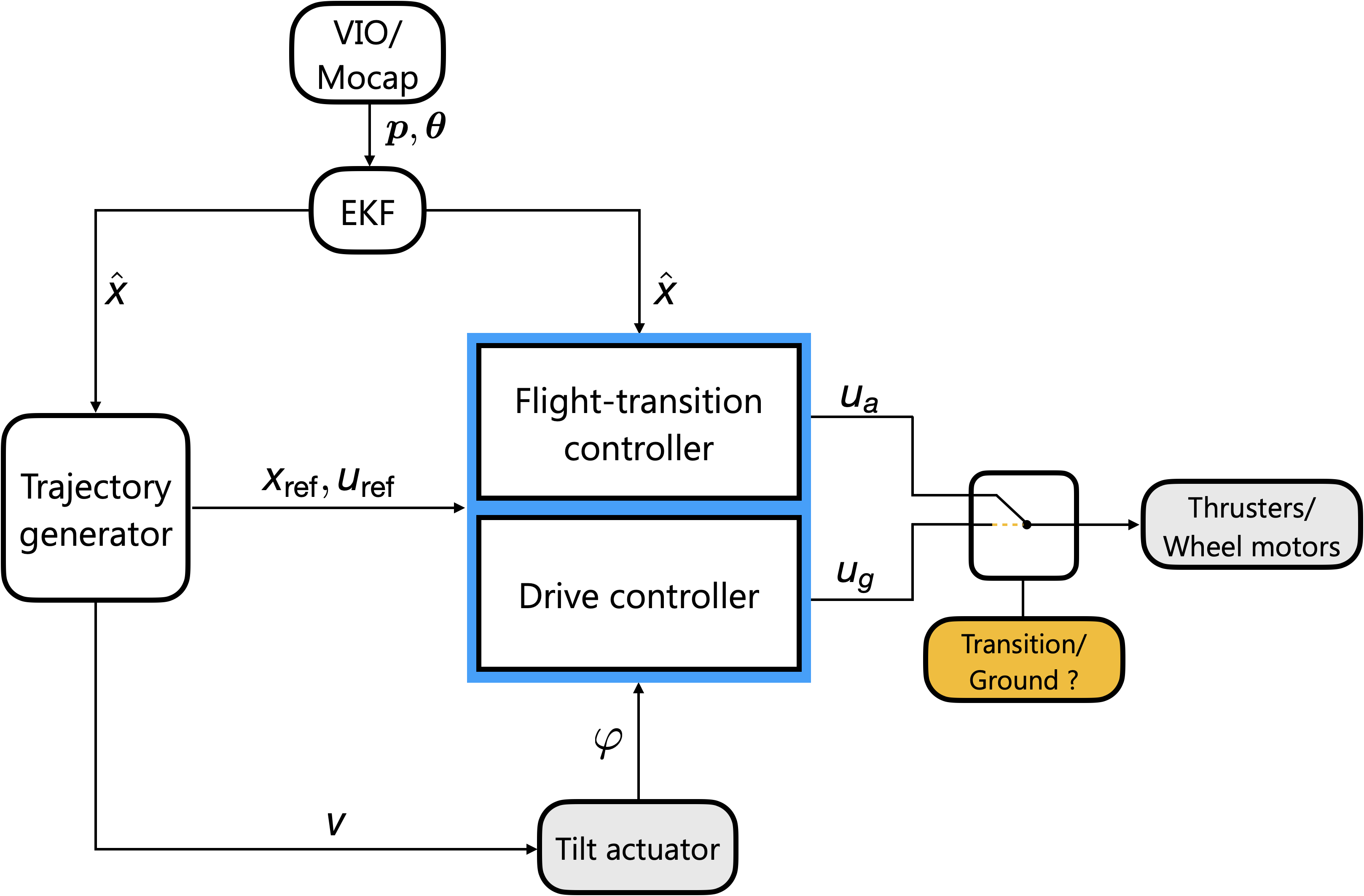}
    \caption{\textbf{Control Architecture.} The control architecture is based on a central module which consists of two controllers: The flight-transition controller, and the drive controller. The control module receives reference points $x_{\textrm{ref}}, u_{\textrm{ref}}$ from a high level trajectory generator, and a position and orientation $\bm p, \bm \theta$ from a motion capture system/VIO system which is fed into the flight controller Extended Kalman Filter (EKF) which provides the state estimate $\hat x$. Using this information the flight-transition controller produces thruster actions $u_a$ and the drive controller produces wheel spin actions $u_g$. Whether the commands are actually sent to the thruster motor electronic speed controllers or wheel motor drivers depends on whether or not a ground contact/transition state have been estimated. The trajectory generator is responsible for producing the tilt velocity $v$ which is fed to the tilt actuator mechanism which in turn produces angle feedback, $\varphi$, to the controller using the mechanism kinematics and encoder count.}
    \label{fig:4}
\end{figure}

\begin{figure*}[t]
    \centering
    \includegraphics[width=1\linewidth]{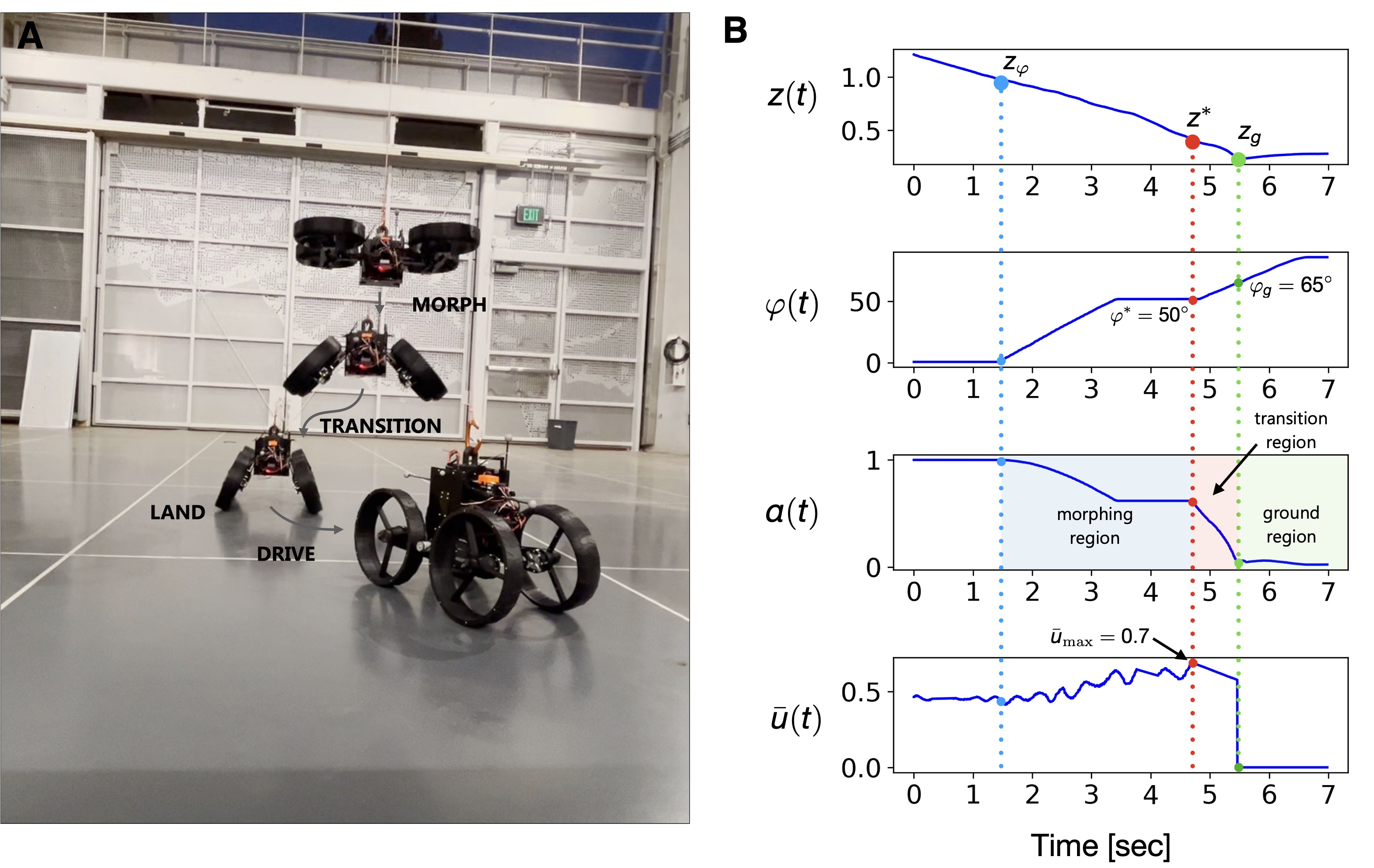}
    \caption{\textbf{Experimental Validation of Dynamic Wheel Landing.} (\textbf{A}) Recorded data during an autonomous wheel landing. In blue $z(t),\varphi(t),\alpha(t),\bar u(t)$, i.e. the altitude, tilt angle, control blending factor, and the mean normalized thrust applied by the four rotors are plotted. The three vertical lines denote the time instances at which the robot begins to morph (height $z_\varphi$), the point at which transition begins (height $z^\ast$) and finally the point at which impact occurs (height $z_g$). The morphing, transition, and ground regions are highlighted in blue red, and green respectively. The morphing height, transition height, and ground contact height ($z_\varphi,z^\ast,z_g$) are indicated along with corresponding angles, and the maximum normalized thrust (\textbf{B}). Snapshots from the performed trajectory.}
    \label{fig:6}
\end{figure*}
Approaching the ground is known to alter the thrust characteristics of conventional drones \cite{MatusVargas2021,SanchezCuevas2017,aich2014analysis,He2020}. Compounded with the fact that the tilted thrusters may introduce unknown aerodynamic interactions, the dynamics of the transition region are largely unknown.

To elucidate how the near-ground aerodynamics affect the wheel landing maneuver, we measured the thrust of the robot as it approached the ground with different tilt angles through static load-cell testing. We also performed smoke-visualization of the flow field at different angles to examine the underlying factors that influence landing stability. The experimental setup and results are summarized in Figure 3. 

The smoke visualization presented in Figure 3(B) reveals that up to $\varphi = 60^\circ$ the interaction of the two rotor flows results in a net downward transfer of momentum since the streams of air combine by mixing and accelerating away from the robot body. Conservation of linear momentum dictates that the net thrust is oriented upwards. In contrast, beyond $\varphi = 60^\circ$, e.g. at $\varphi = 70^\circ$, this trend reverses. The interaction of the two air jets becomes unstable resulting in a significant portion of the flow being re-oriented in the undesired direction. This reversal results in a net loss of thrust due to the loss of downward fluid momentum. Furthermore, the instability of the air jet interaction, which can be seen in Supplementary Videos 7 and 8, also cause a greater time variability of the thrust, increasing system dynamic uncertainty.

We quantified this phenomenon by taking ground effect measurements, i.e. the percent of additional thrust gained due to ground proximity, by mounting ATMO on a robot arm and taking force measurements using a load cell. Our results, shown in Figure 3, demonstrate a significant ground effect for angles $\varphi=40^\circ,50^\circ,60^\circ$, evidenced by the increasing thrust as the robot approached the ground. The largest ground effect was found at $\varphi=50^\circ$ with an increase of thrust of almost $20\%$. As expected by the smoke visualization, the trend of increased thrust with ground effect did not persist at larger tilt angles. For $\varphi=70^\circ$, approaching the ground resulted in a loss of thrust, i.e. an aerodynamic suction. Entering this state is dangerous since the loss of thrust will result in a larger than anticipated impact velocity. Furthermore, the time variability of the thrust was much higher for $\varphi=70^\circ$ compared to the other angles as seen by observing the increase in area of the shaded portion in Figure 3(C) for $\varphi=70^\circ$. Taking into account the standard deviation, we observe that the thrust force can fall to as low as $85\%$ of the thrust away from the ground as the robot approaches the ground for $\varphi=70^\circ$. 

\subsection*{Ground-Aerial Morpho-Transition Control}

There are three distinct phases in a dynamic wheel landing: 1) quadrotor flight, 2) morphing flight, and 3)  near ground morpho-transition. These phases are depicted in Figure 1(A). The robot is initially in flight at a height $z_0$, begins morphing at some height $z_\varphi$, and enters the final near-ground transition phase with a tilt angle $\varphi^\ast$ at a height $z^\ast$ above the ground.

During morphing flight, the thruster tilting causes a dynamic coupling between roll and horizontal motion. This is illustrated in the morphing flight force diagram of Figure 1(A). If ATMO attempts to compensate for a disturbance by applying a roll torque, a horizontal component of the thrust force is also produced in the undesired direction. As a result, rolling also leads to a translation in the opposite direction, as opposed to the motion of conventional quadrotors. Thus, off-the-shelf flight control architectures which rely on cascaded PID loops that decouple attitude and position control do not directly apply for the stabilization of mid-flight morphing robots of this type, as shown via numerical experiments presented in Supplementary Note 1 and Video 10.

In the near-ground transition phase the system dynamics become even more complex due to ground-proximity aerodynamics and air jet interaction. Furthermore, landing on wheels with the largest possible tilt angle requires approaching or surpassing the critical angle. In this phase, position and velocity control are lost when attempting to compensate for gravity, due to actuator saturation. The only way to maintain attitude control is by foregoing vertical position and velocity control, making this the most challenging phase of the maneuver.

To take into account the different phases of the dynamics, we designed a custom model-predictive controller with a time-varying cost function that adapts to mid-flight change of body posture and ground proximity (see Methods). Using a model of morphing flight dynamics, actuator limitations, and taking into account the influence of ground proximity on thrust performance, our model-predictive control scheme stabilizes ATMO in flight across all possible body posture configurations. This results in a time and shape varying behavior that adapts to the proximity to the ground, as well as the degree of morphing, thus linking all three phases of the wheel landing task together. 

To achieve this we use specialized objective functions for the flight, morphing, and morpho-transition phases of the maneuver. These are changed online in the optimization-based controller using a blending factor $\alpha(z,\varphi)$ which depends on the current robot altitude $z(t)$ and body tilt angle $\varphi(t)$. The blending factor smoothly switches the cost function for flight to a transition cost function as soon as ATMO enters the transition phase of the maneuver. 

The transition was set to begin at $z^\ast = 0.45 \ \textrm{m}$ above the ground. At this height, ground effect offers additional thrust, as evidenced by the morpho-transition aerodynamics, slowing down impact and enabling greater tilt angles at landing. Indeed, as described previously, ground effect has a net positive effect during landing up until $\varphi_{\max}^\ast = 70^\circ$. By limiting the tilt angle to this value, we ensure that ATMO is always assisted by the additional thrust from the ground effect during morpho-transition, allowing it to land and takeoff with body postures greater than the critical angle. The use of ground effect during the landing is a key reason for which gradual transformation may be advantageous compared to rapid transformation in morpho-transition landings.

The controller takes advantage of the aerodynamics by going effectively open loop in vertical position and velocity as soon as ATMO reaches the transition region. By virtue of the additional thrust gained by the aerodynamic effect, this enables the controller to prioritize attitude stabilization and land safely at the desired impact velocity, despite the inherent tradeoff between attitude stabilization and impact velocity at past-critical tilt angles. For more details on the choice of objective function, the reader is referred to the Methods section.

Finally, to enable smooth driving as soon as the robot touches the ground, the wheel motors are activated shortly before impact. Here, the drive controller begins sending wheel spin commands in order to track the reference position. Once the robot touches the ground, the thrust commands no longer reach the thrust motors. The actuator switching logic is implemented with a state machine that uses the current system state to determine whether the robot is grounded or not, and cuts or allows the signal flow to the wheel motors accordingly. For a full description of the controller the reader is referred to the Methods section. The overall control architecture is summarized in Figure 4, and the algorithm details are given in the Methods section. 

\begin{figure*}[!t]
    \centering
    \includegraphics[width=1\linewidth]{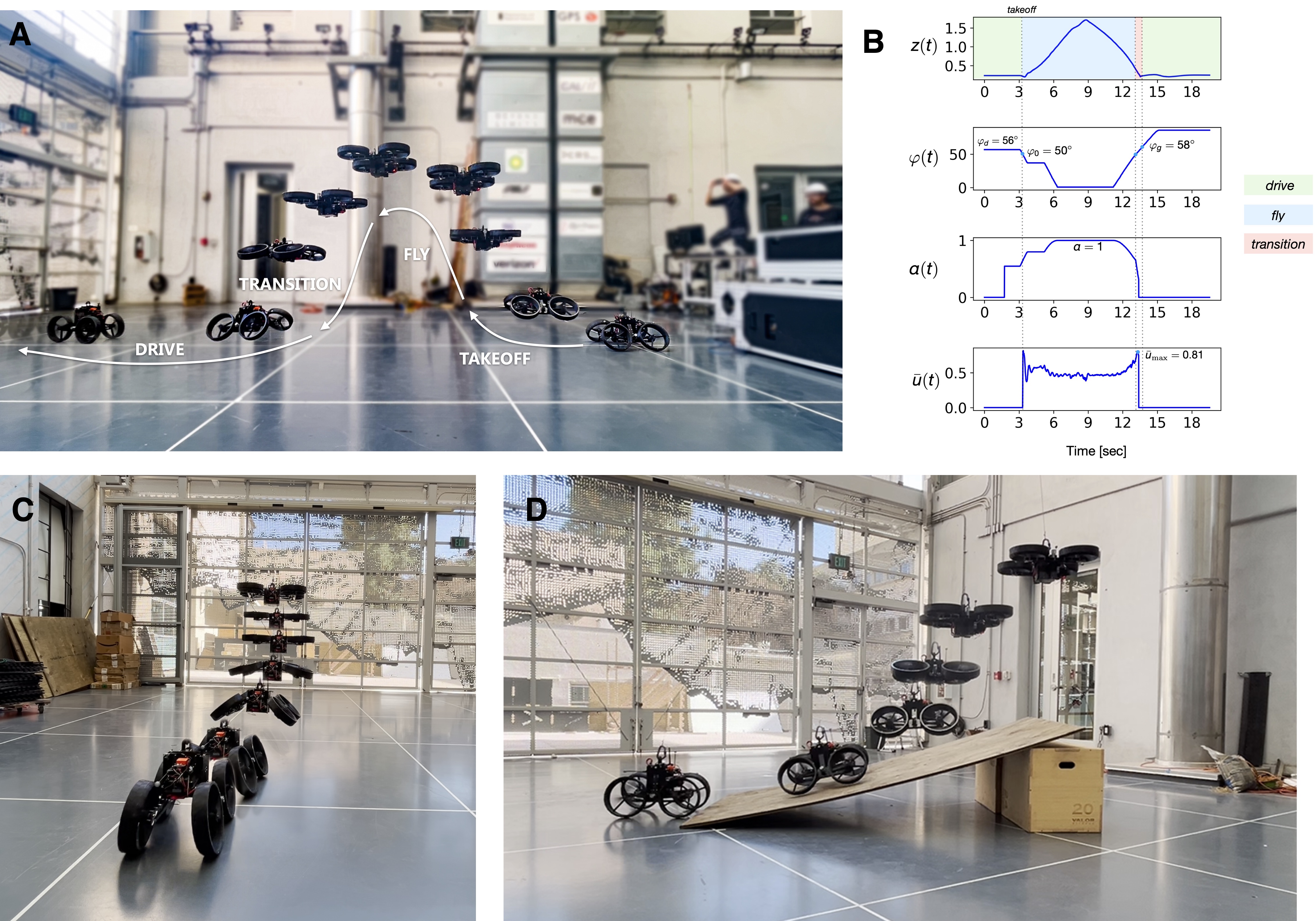}
    \caption{\textbf{Experimental Validation of Ground-Aerial Transition Maneuvers. }(\textbf{A}) Flight Initiating Driving Takeoff. Robot moving from right to left begins in driving configuration, drives and jumps up by switching on its thrusters. It transforms to quadrotor configuration until it reaches the apex of the trajectory, and finally it begins its descent with forward velocity, transforming into drive mode, smoothly landing, and continuing its driving motion (Video 3 in Supplementary Materials). (\textbf{B}) In blue $z(t),\varphi(t),\alpha(t),\bar u(t)$, i.e. the altitude, tilt angle, time varying mixing controller coefficient, and the average normalized thrust. The driving, flying, and transitioning modes are highlighted in  green, blue, and red respectively. (\textbf{C}). Example of landing with a forward velocity. In this example the controller was further tuned to improve disturbance rejection performance. (\textbf{D}) Example of using the proposed controller to land on a 25 degree slope and continue driving.}
    \label{fig:6}
\end{figure*}

\subsection*{Experimental Validation}

The ground-aerial morpho-transition controller was applied for a dynamic wheel landing in the Caltech Center for Autonomous Systems and Technology (CAST) flight arena using a motion capture system to facilitate state estimation (see Methods section). Snapshots from the resulting experiment are shown in Figure 5(A) and a video of the maneuver can be found in the supplementary materials (Video 2). In this experiment the controller was used to track a reference trajectory in space that consisted of a descent with some forward motion while tilting the wheel-thrusters, landing on the wheels, and finally driving forward.

As can be seen labeled in Figure 5(B) on the $\varphi(t)$ graph, the final tilt angle at landing is $\varphi_g=65^\circ$, showing that the robot is successfully able to land at a tilt angle that is past the critical angle. This is achieved due to the change of cost function in the transition phase evidenced by the sharp drop of $\alpha(t)$. By virtue of the changing cost function the robot can continue tilting its wheel-thrusters while still maintaining the desired attitude by reducing the emphasis on holding vertical position or velocity and prioritizing only attitude stabilization. Ultimately, the tilt angle at landing is $\Delta\varphi = 15^\circ$ above the maximum in-flight tilt angle, while the maximum mean normalized thrust during the maneuver is $\bar u = 0.7$. We hypothesize that this is possible in part due to the ground effect that was measured in Figure 3(C). Note that, the robot shows a noticeable lateral drift during the maneuver (Video 2). This might have been due to wind disturbances present in the CAST flight arena or controller tuning. We repeated the experiment with different cost function parameters (see Methods section) and report noticeably less lateral drifting, as can be seen in Video 4 of the supplementary materials. The safety tether was also removed showing that additional forces were not applied to the robot during the maneuver. 

We also validated our control method by performing a driving takeoff followed by a dynamic wheel landing. In this maneuver the robot is tasked with taking off while driving, dynamically transitioning to flight by deploying as a quadrotor, and finally morphing again to land on wheels and continue driving. The resulting maneuver is depicted in Figure 6(A), and the evolution of the key states, and parameters are graphed in Figure 6(B). A video of the maneuver can be found in Supplementary Materials, Video 3. The robot started in drive mode with $\varphi_d =56^\circ$ tilt angle in driving configuration. As the robot drove forward, the throttle was increased, triggering the robot to decrease the tilt angle. At $\varphi_0 = 50^\circ$, the thrusters were activated and the throttle increased, thus propelling the robot into the air. After some moments in the air, the robot descended and smoothly transitioned into driving configuration. Once in the transition phase $z<z^\ast$, the wheels were activated and the robot smoothly continued its forward driving motion. As can be seen in Figure 6 (B), the maximum tilt angle while in flight was $50^\circ$ to avoid issues with actuator saturation, and the final tilt angle at landing was $\varphi_g = 58^\circ$. The maximum mean normalized thrust in the landing phase of the maneuver was $\bar u = 0.81$. This is significantly higher than in the previous maneuver potentially due to the more aggressive forward motion which required higher actuator action to produce a pitch moment and counteract gravity simultaneously. The repeatability of landing with forward velocity and continuously driving was confirmed by further experimentation. An example maneuver is shown in Figure 6(C) and Video 6. 

Finally, we showcased an important use case of mid-air transformation by demonstrating a dynamic wheel landing on an inclined plane. In the maneuver depicted in Figure 6(D), and shown in Video 5, ATMO was able to land smoothly on a slope of known height and position and continue driving. Indeed landing on slopes might be dangerous for conventional quad-rotors due to the potential of toppling over. Transforming prior to landing may help to avoid this.

\FloatBarrier
\section*{Discussion}

We have shown in this study how mid-air transformation can be used to achieve dynamic ground-aerial transition maneuvers. Our results demonstrate the possibility of a robot starting in quadrotor flight mode and smoothly transforming to land on its thrusters which are re-purposed as wheels. We also demonstrated the inverse maneuver which consists of rapid takeoff combined with forward driving locomotion, as well as landing on a slope. 

To achieve this, we first developed a specialized robot, ATMO, that contends with mid-air thrust forces to transform while flying using a {\bl specialized morphing mechanism}. ATMO distinguishes itself from other flying-driving robots by virtue of a self-locking tilt actuator mechanism that enables mid-air transformation with minimal actuation requirements, lower cost, and simpler overall design. Although other morphing quadrotor designs have been employed for fitting through narrow gaps \cite{Falanga2019,zhao2017,desbiez2017,bucki2019,bucki2021}, or for achieving full actuation or manipulation capabilities \cite{zheng2020, Li2024, Ryll2022}, fewer works have used mid-air transformation as a means to enhance ground-aerial locomotion performance \cite{Zhao2023}. 

Secondly, we developed a model-based control scheme covering the full operational envelope of flying, driving and transitioning. To tackle the problem of actuator saturation that occurs as the robot tilts its thrusters to land on its wheels we used a decomposition of the control objective function into a convex combination of specialized objective functions for each locomotion mode. This approach offers a flexible framework for controlling mid-air transforming robotic systems in ground-aerial transition. Since the nature of the morpho-transition problem for ATMO is one where actuator constraints play a major role, using a model predictive controller allows us to seamlessly incorporate constraints such as the thrust limits into the controller. While it is in theory possible to use different control methods such as gain scheduling combined with linear quadratic control, these methods require significant engineering effort and offer less interpretability.  

Furthermore, we were able to show that the developed controller can enable landings with tilt angles past the actuator saturation limits. High tilt angles at landing are an important metric for controller performance, since they are challenging to achieve with standard control methods that do not account for actuator constraints. Landing with larger tilt angles also enables ATMO to clear {\bl uneven terrain} that may be present at the landing site. Of course, since flying with a larger tilt angle also demands a higher level of thrust, it is less energetically beneficial. As a result, landing with lower, or no, tilt angle may be acceptable and consume less power in some cases - for example when the landing site is flat and ground transformation is possible. 

Third, we tested the aerodynamics of the near-ground morpho-transition phase and found that that the flow regime of four tilted, interacting rotors approaching the ground differs significantly from that of non-transforming, vertically descending quadrotors. We showed that the ground-effect relation persisted for tilted rotors up to and including the $\varphi = 60^\circ$ tilt angle case and then reversed due to fluid dynamic instability. We were able to use this effect to achieve landings past the critical actuator saturation angle. This result suggests that formally understanding and utilizing the complex aerodynamic interactions of morphing flight may be an effective way to increase the agility of transforming ground-aerial robotic systems.

Although the literature on quadrotor ground effect is expansive and significant progress has been made in aerodynamic characterization and analytical models \cite{SanchezCuevas2017}, aerodynamic models for robotic systems with tilted thrusters approaching the ground are still in their infancy \cite{Soldado2024,GarofanoSoldado2022}, even more so when rotor-rotor interactions are included in the analysis. Our findings may be applied for emerging scenarios such as passive wall tracking using tilted propellers \cite{Ding2022}, collision-free reactive navigation \cite{Ding2023}, or energetically efficient perching maneuvers that exploit proximity effects like the ceiling effect \cite{Hsiao2018,Hsiao2023} but have not yet considered the potential benefits of using tilted rotors \cite{LiuYiliang2024}.

Overall, while dynamic transition maneuvers were successfully demonstrated experimentally, the conditions were controlled to facilitate rapid development. A key simplification was the use of a motion capture camera system for accurate and fast estimation of position and orientation of the robotic system which exceeds what can be obtained by current onboard sensors like satellite or inertial measurement units. Future investigation is warranted to examine the extent to which the dynamic transition maneuvers presented here extend to real-world scenarios in which robots may face complex, unstructured terrain and must take decisions based on partial sensor information subject to noise. Immediate next steps involve incorporating a vision-based method which autonomously decides on optimal landing locations and the optimal landing body configuration and extending ATMO's sensor suite to enable flight transitions in outdoor environments. With the capabilities that have already been showcased in this paper as well as some of these improvements, mid-air transforming robotic systems like ATMO promise to provide real-world assistance in scenarios where transforming before touch-down or take-off protects mission-critical hardware and actuators from the dangers of hostile terrain while enabling increased agility. 

\FloatBarrier
\section*{Methods}

\subsection*{Controller Description}

The control architecture is summarized in Figure 4. Here, $\bm x$ is the system state, $\bm {u_a}$ are the thruster/aerial control inputs and $\bm {u_g}$ are the ground/driving control inputs while $\bm x_\textrm{ref},\bm u_\textrm{ref}$ denote the state and input references to be tracked. The controller is based around a central module consisting of two separate controllers: (1) the flight-transition controller which handles the aerial control of the robot and outputs the four thruster commands $\bm {u_a}$ and (2) the drive controller which handles ground control and outputs the wheel motor commands $\bm {u_g}$. As shown in Figure 4, the flight-transition controller receives the desired reference $x_\textrm{ref},u_\textrm{ref}$ and a state estimate $\hat x$ and stabilizes the robotic system along the reference trajectory by solving the following optimal control problem in a receding horizon control fashion:
\begin{mini*}|1|[2]
        {\bm x,\bm u}
        {\int_0^{t_f} \left[\alpha(\bm x) L_1(\bm x,\bm u) + (1-\alpha(\bm x)) L_2(\bm x,\bm u) \right]d t}
        {\label{eq:cost}}{}
        \addConstraint{}{\bm {\dot{x}} =  \bm f(\bm x,\bm u, \varphi),}{\quad \forall t\in[0,t_f]}
        \addConstraint{0 \le u_a^i \le 1, \quad }{}{\forall i=\{1,\dots,4\}. }
    \end{mini*}
Here, $L_1(\bm x,\bm u)$ and $L_2(\bm x,\bm u)$ are the cost functions associated with flying and transitioning, $t_f$ is the horizon of the optimization, and $u_a^i$ are the normalized thrust values of each thruster, constrained between 0 and 1 to account for actuator saturation. The system dynamics $\bm {\dot x} = \bm f(\bm x,\bm u,\varphi)$ are enforced by numerically discretizing the optimal control problem in a multiple shooting scheme \cite{Bock1984}. $N$ collocation points of time $\delta t$ over a time horizon $t_f$ are used. This leads to a nonlinear program,
\begin{mini*}|1|[2]
    {\bm x[\cdot],\bm u[\cdot]}
    {\sum_{k=0}^{N-1} L(\bm x[k],\bm u[k])}
    {\label{eq:cost}}{}
    \addConstraint{\bm x[0]}{=\bm x_0}{}
    \addConstraint{\bm x[k+1]}{=\bm f_{\textrm{RK4}}(\bm x[k],\bm u[k],\delta t),\quad}{\forall k}
    \addConstraint{0 \le u_k \le 1. \quad }{}{\forall k }
\end{mini*}
\noindent where $\bm f_{\textrm{RK4}}$ is a Runge-Kutta integrator of the dynamics function $\bm f$. To leverage the capabilities of new embedded solvers the nonlinear program is solved using sequential quadratic programming. This is implemented on the onboard computer using software packages CasADi and ACADOS \cite{Verschueren2019acadosaMO}. By using the SQP approach executed in a real time iteration scheme the model predictive controller is run at 150Hz on the onboard computer. The horizon time is chosen as $t_f = 1.0$ second and $N=10$ collocation points are used. 

\subsection*{Objective Function}
The objective function used in the optimization-based controller consists of two quadratic cost terms $L_1,L_2$ which represent the flight and transition phases respectively. The cost is varied according to the tilt angle and the altitude with the blending factor $\alpha(z,\varphi)$,
\begin{align*}
    \alpha(z,\varphi) &= f(z)\cos\varphi, \\
   f(z) &=
    \begin{cases} 
    1.0 & \text{if } z \geq z^*, \\
    \frac{z - z_g}{z^* - z_g} & \text{if } z_g \leq z \leq z^*, \\
    0.0 & \text{otherwise,}
    \end{cases}
\end{align*}
\noindent where $z_g$ is the ground height. The blending factor is proportional to two terms. The first term $\cos\varphi$ decays to zero as $\varphi\rightarrow 90^\circ$, and the second term $f(z)$, active only in the transition region, ensures that $\alpha\rightarrow 0$ always as the robot touches the ground. The cost function is updated once every 10 iterations of the model predictive control scheme. This ensures minimal computational overhead, as well as giving sufficient time for the optimizer to warm up the solution. 

$L_1,L_2$ are chosen to be quadratic costs as follows:
\begin{align*}
L_j &= ||\bm x - \bm x_{\textrm{ref}}||^2_{Q_j} + ||\bm u - \bm u_{\textrm{ref}}||^2_{R_j}, \ \forall j\in\{1,2\},
\end{align*}
The cost matrices associated to each flight phase were chosen as,
\begin{align*}
    Q_1 &= \textrm{diag}(1,1,1,10,10,20,0.1,0.1,0.1,3,5,1.5), \\
    Q_2 &= \textrm{diag}(0,0,0,0,0,0,0,0,0,3,5,1.5), \\
    R_1 &= R_2 = \textrm{diag}(.1,.1,.1,.1),
\end{align*}
for the experiment shown in Figure 5 and Video 2. For the remainder of the experiments, the cost matrices were adjusted to improve controller performance and were chosen as:
\begin{align*}
    Q_1 &= \textrm{diag}(1,1,1,10,10,18,0.1,0.1,0.8,1.5,2.7,2.0), \\
    Q_2 &= \textrm{diag}(0,0,0,0,0,0,0,0,0,1.5,2.7,2.0), \\
    R_1 &= R_2 = \textrm{diag}(.1,.1,.1,.1),
\end{align*}

\noindent Note that the cost-matrix, $Q_2$, associated with the transition phase reflects the loss of controllability in $x,y,z,\dot x,\dot y, \dot z$ and the increased priority on angular velocities.

\subsection*{Trajectory generator}
The trajectory generator outputs the desired state and input trajectory and sends the reference points in real time to the robot through the ROS2 network. For the experiments conducted in this work manual control commands were supplied for the 3D position of the robot as, $\bm p ^{\textrm{ref}}(t) = \bm p(t) + \bm v T_s$, where $T_s$ is the sampling time of the controller. The velocity commands $\bm v = (v_x,v_y,v_z)$ are received from a radio controller and limited to $1 \ \textrm{m}/\textrm{s}$ for safety. The attitude and angular velocity reference states are set to zero throughout the experiment. 

The tilt angle velocity reference is assigned according to the phases of the dynamic wheel landing task. The important parameters are the height above the ground, $z$, and whether the robot is grounded or not (grounded: $\lambda=1$, not grounded: $\lambda=0$). Additionally, if the robot is on the ground the user can indicate that the robot should take off by setting the overall throttle $c$ to be greater than $50\%$ throttle. This logic is encoded overall as the following tilt velocity reference:
\begin{equation*}
\dot \varphi_{\textrm{ref}}(z,c,\lambda) =
    \begin{cases}
         v_{\max} & z < z_\varphi, \ \lambda=0,\\
         -v_{\max} & c \ge 0.5 , \ \lambda=1,\\
         0 & \textrm{otherwise}.
    \end{cases}
\end{equation*}

\subsection*{Kinematic Modeling}
The robot can change its posture through a closed kinematic chain that is actuated by a brushed DC motor with a relative encoder for sensing. Using the known pivot point $D$, the kinematics of the closed linkage can be evaluated by solving the following system for $\theta,\varphi$ as a function of the linear displacement of the first link $x = ||OA||$,
\begin{equation*}
    \begin{bmatrix}
    D_{x}\\D_{y}
    \end{bmatrix} =
    \begin{bmatrix}
    h\\x
    \end{bmatrix} +     \begin{bmatrix}
    \cos\theta & \cos\varphi\\
    \sin\theta & -\sin\varphi
    \end{bmatrix}
    \begin{bmatrix}
    d_1\\d_2
    \end{bmatrix},
\end{equation*}
\noindent where $\theta$ is an internal mechanism angle, $\varphi$ is the tilt angle, and $h=||AB||$, $d_1=||BC||$, and $d_2=||CD||$ are the mechanism link lengths as depicted in Figure 2(B). To relate the linear displacement to the number of revolutions that the motor should carry out, we use the pitch thread $\alpha=0.8 \ \textrm{cm}$ and the number of encoder counts per revolution of the tilt motor output shaft $n=1632.67$ for the Pololu high power 34:1 motor. The equations were solved numerically for the tilt angle ($\varphi$) to characterize the tilt angle against encoder count (where the encoder is set to zero in drone configuration). The analytical results for the tilt angle were compared to measured tilt angle values at various encoder counts on the experimental platform (see supplementary Figure S2). The good agreement validates the kinematic equations for tracking angle. The parameters on our robotic system were: $h=1.6 \ \textrm{cm}$, $d_1 = 5.2 \ \textrm{cm}$, $d_2 = 4.6 \ \textrm{cm}$, $D_{x} = 6.8 \ \textrm{cm}$, $D_{y} =5.1 \ \textrm{cm}$.

\subsection*{Dynamic Modeling}
We model the robot as a set of 7 inertial components: the robot base, the left arm and right arm, and the four spinning propellers. The wheels are considered fixed since the spinning motion of the wheels is unused for stabilization during the flight drive transition. 

To obtain the system equations of motion the Lagrangian approach is employed. First the generalized coordinates and velocities are defined as: $\bm q = (\bm p,\bm \theta)$ and $\bm u =(\bm v, \bm \omega)$ where $\bm p$ is the position of the robot center of mass, $\bm \theta$ is the orientation of the robot relative to the inertial world frame (expressed as a vector of Euler $zyx$ angles), $\bm v$ is the velocity of the center of mass, and $\bm \omega$ is the angular velocity of the robot relative to the world frame expressed in the body frame. 

The control inputs are $\bm \tau = (u_1,u_2,u_3,u_4)$, where $u_i, i\in\{1,...,4\}$ denote the normalized propeller RPMs. The thrust and moment due to the transfer of momentum caused by the rotating propellers pushing air through the rotor disk is calculated by assuming a linear relationship between the normalized propeller RPMs and the thrust/moment generated by each thruster-propeller combination i.e. $T_i = k_T u_i$ and $M_i = k_M k_T u_i$. 

The control inputs $u_i$ are thus normalized between $\left[0,1\right]$ meaning that the max thrust of each propeller-thruster combination is $T_{\max} = k_T$. The thrust and moment are assumed to act parallel to the rotation axis of the propeller. The parameters $k_T,k_M$ are identified experimentally as detailed in the system identification section in Supplementary Note 3. To derive the system dynamics equations the projected Newton-Euler method implemented in the open-source MATLAB package proNEu \cite{Hutter2011} is employed. This results in equations of motion of the following form,\begin{align*}
    \bm M(\bm q,\varphi)\bm{\dot u} + \bm b(\bm q,\bm u,\varphi) + \bm g(\bm q,\varphi) &= \bm S(\varphi) \bm \tau, \\
    \dot \varphi &= v.
\end{align*}
Here $\bm M(\bm q,\varphi)$ denotes the mass matrix, $\bm b(\bm q, \bm u,\varphi)$ encodes the coriolis terms, and $\bm g (\bm q,\varphi)$ encodes the gravitational dynamics. The actuation matrix $\bm S(\varphi)$ maps the generalized control input into the generalized acceleration space. 
These dynamics are then written in state-space form,
\begin{align*}
    \bm{\dot{x}} &= \bm f(\bm x,\bm u, \varphi), \\
    \bm x &=(\underbrace{x,y,z}_{\bm p},\underbrace{\theta_z,\theta_y,\theta_x}_{\bm \theta}, \underbrace{v_x,v_y,v_z}_{\bm v},\underbrace{\omega_x,\omega_y,\omega_z}_{\bm \omega})\in \mathbb{R}^{12}, \\
    \bm u &= (u_1,u_2,u_3,u_4)\in\mathbb{R}^4.
\end{align*}
A particularity of our approach is that the tilt angle $\varphi$ is modeled as a pure integrator, reflecting the physical properties of the non-compliant, self-locking morphing mechanism. This requires $\varphi$ to be a kinematic parameter rather than a dynamics variable. Note that all aerodynamic effects due to approaching the ground, walls, or other obstacles as well as the unknown effect of the interacting rotor flows are neglected due to difficulty in producing simple analytical models of these phenomena over the desired range of operating conditions. 

\subsubsection*{Critical angle}
The critical angle is an important parameter for morphing flight dynamics for robots with thrusters that tilt away from the vertical plane. It is calculated as the point where the weight of the robot equals the maximum projection of the thrust in the vertical axis: $\cos\varphi_c = \frac{m g}{T_{\max}}$. As mentioned in the results section, the in-flight tilt angle is limited to $\varphi_{\max} = 50^\circ$. This value is chosen to ensure that $35 \%$ of thrust is available for disturbance rejection i.e. the thrust at the maximum tilt angle should be 1.35 times greater than the weight of the robot:
\begin{equation*}
    \frac{T_{\max} \cos\varphi_{\max}}{m g} = 1.35.
\end{equation*}
For a thrust to weight ratio of 2.1, $T_{\max}/m g = 2.1$ then $\varphi_{\max}=50^\circ$.
\subsection*{Tilt and driving control}
The tilt angle is controlled by receiving desired tilt angle velocity commands $v$ directly from the trajectory generator. The tilt angle $\varphi$ is sensed using the mechanism kinematics and sent back to the controller. It is also used to limit the tilt angle to $\varphi \le \varphi_{\max} = 50^\circ$ in flight and $\varphi \le \varphi_{\max} = 70^\circ$ in the transition phase.

For driving we use a model of a differential drive vehicle with control inputs $u_g = (\dot \phi_l,\dot \phi_r)$, i.e., the wheel rotation rates of the left and right wheels. This is supplied on the hardware side by sending a PWM signal to the drive motors. The dynamics of the ground position and orientation $(x,y,\theta)$ due to the wheel spinning are given by
\begin{align*}
    \dot x &= \frac{1}{2} R (\dot \phi_l + \dot \phi_r) \cos\theta \equiv V \cos\theta,\\
    \dot y &= \frac{1}{2} R (\dot \phi_l + \dot \phi_r) \sin\theta \equiv V \sin\theta,\\
    \dot\theta &= \frac{R}{2 l} (\dot \phi_l - \dot \phi_r) \equiv \omega,
\end{align*}
with $V = \frac{1}{2} R (\dot \phi_l + \dot \phi_r) $ and $\omega = \frac{R}{2 l} (\dot \phi_l - \dot \phi_r)$. $R$ and $l$ denote the wheel radius and half wheel base respectively. Given a desired ground velocity $(V,\omega)$ we calculate the associated wheel speeds as $\dot\phi_r = \frac{1}{R}(V - l \omega)$ and $\dot\phi_l = \frac{1}{R}(V + l\omega)$.

\subsection*{Aerodynamic testing}

The experimental setup shown in Figure 3 was used to test the thrust characteristics as the robot approaches the ground with different tilt angles. The load cell used consisted of a 1-axis S-type unit procured from Interface that was calibrated against precise calibration weights. We ensured the load cell was aligned with the direction of gravity prior to experiments. The height of the robot to the ground was streamed in real time using the OptiTrack measurement system and the angle was computed from the encoder counts using the onboard kinematics calculations. For each data point the thrusters were set to $50\%$ thrust and the thrust, $T$, measured by the S-type load cell was recorded for $t_f = 15$ second intervals to produce the set $\{T(t,\varphi, z) | \ t\in [0,t_f]\}$ for each $\varphi$ and $z$ considered. After each trial the average thrust $\bar T(z,\varphi)$ was recorded, as well as the standard deviation of the thrust (in time) during the trial. The normalized thrust values reported in Figure 3 were obtained as $\frac{\bar T(z,\varphi)}{\bar T_\infty(\varphi)}$ where $\bar T_\infty(\varphi)$ is the average thrust measured at the load cell far from the ground. In order to maintain the thrust level across experiments steady, a 24V power supply was used instead of the onboard 6S LiPo battery. For the unnormalized thrust values obtained in this experiment, the reader is referred to Supplementary Note 2. 

The smoke visualization was done using a laser placed on the flat ground shining onto a $45^\circ$ mirror in order to create a laser sheet in the plane of the front two robot thrusters. The smoke was supplied using standard fog machines into one of the middle plane between the front two thrusters. The fog machines were held on the robot axis of symmetry between the two front thrusters to ensure that smoke was entrained into both sets of propellers. The movement of the smoke was observed over all possible tilt angles as seen in Supplementary Videos 4 and 5. 

\section*{Data availability}

Data will be provided upon request.

\section*{Code availability}

Code will be provided upon request.

\bibliographystyle{ieeetr}
\bibliography{main.bib}

\section*{Acknowledgements}
We are grateful to Gabriel Margaria, Quentin Delfosse, Vincent Gherold, Simon Gmür and Alejandro Stefan-Zavala who helped with flight experiments and Scott Bollt and Julian Humml who provided their assistance in setting up the lasers for smoke visualization and the robotic arm used for load cell testing. We also thank Alexandros Rosakis for comments on an early draft of the manuscript. This project was supported by funding from the Center for Autonomous Systems and Technologies at Caltech. Ioannis Mandralis is an Onassis Scholar and was supported by the GALCIT endowment graduate student fellowship. 

\section*{Author contributions}
All authors designed research and were involved in discussions to interpret the results; I.M. wrote software and performed control experiments; I.M. designed the control architecture with input from R.M.M., A.R., and M.G.; I.M. performed load cell tests with input from M.G.; I.M. and M.G. performed smoke visualization experiments; I.M., R.N., and M.G. came up with the robot design concept and R.N. manufactured initial prototype; I.M. drafted the paper, and all authors helped edit and review.

\section*{Competing interests}
The authors declare they have no competing interests. 

\end{document}